\documentclass[sigconf,table, prologue]{acmart}
\AtBeginDocument{%
  }

\setcopyright{acmlicensed}
\copyrightyear{2025}
\acmYear{2025}
\acmDOI{XXXXXXX.XXXXXXX}
\acmConference[]{}

\usepackage{graphicx}
\usepackage{subcaption}
\usepackage{booktabs}
\usepackage{makecell}
\usepackage{pifont} 
\usepackage{tabularx}
\usepackage{lscape}
\usepackage{booktabs}
\usepackage{makecell}
\usepackage{multirow}

\begin{document}

\title{POIFormer: A Transformer-Based Framework for Accurate and Scalable Point-of-Interest Attribution}

\author{Nripsuta Ani Saxena}
\email{nsaxena@usc.edu}
\affiliation{%
\institution{University of Southern California}
\city{Los Angeles}
\state{California}
\country{USA}
}

\author{Shang-Ling Hsu}
\email{hsushang@usc.edu}
\affiliation{%
\institution{University of Southern California}
\city{Los Angeles}
\state{California}
\country{USA}
}

\author{Mehul Shetty}
\authornote{Both authors contributed equally to this research.}
\email{mehulshe@usc.edu}
\affiliation{%
\institution{University of Southern California}
\city{Los Angeles}
\state{California}
\country{USA}
}

\author{Omar Alkhadra}
\authornotemark[1]
\email{alkhadra@usc.edu}
\affiliation{%
\institution{University of Southern California}
\city{Los Angeles}
\state{California}
\country{USA}
}

\author{Cyrus Shahabi}
\email{shahabi@usc.edu}
\affiliation{%
\institution{University of Southern California}
\city{Los Angeles}
\state{California}
\country{USA}
}

\author{Abigail L. Horn}
\email{hornabig@usc.edu}
\affiliation{%
\institution{University of Southern California}
\city{Los Angeles}
\state{California}
\country{USA}
}
  

\begin{abstract}
Accurately attributing user visits to specific Points of Interest (POIs) is a foundational task for mobility analytics, personalized services, marketing and urban planning. However, POI attribution remains challenging due to GPS inaccuracies, typically ranging from 2 to 20 meters in real-world settings, and the high spatial density of POIs in urban environments, where multiple venues can coexist within a small radius (e.g., over 50 POIs within a 100-meter radius in dense city centers). Relying on proximity is therefore often insufficient for determining which POI was actually visited.
We introduce \textsf{POIFormer}, a novel Transformer-based framework for accurate and efficient POI attribution. Unlike prior approaches that rely on limited spatiotemporal, contextual, or behavioral features, \textsf{POIFormer} jointly models a rich set of signals, including spatial proximity, visit timing and duration, contextual features from POI semantics, and behavioral features from user mobility and aggregated crowd behavior patterns--using the Transformer's self-attention mechanism to jointly model complex interactions across these dimensions. By leveraging the Transformer to model a user's past and future visits (with the current visit masked) and incorporating crowd-level behavioral patterns through pre-computed KDEs, \textsf{POIFormer} enables accurate, efficient attribution in large, noisy mobility datasets.
Its architecture supports generalization across diverse data sources and geographic contexts while avoiding reliance on hard-to-access or unavailable data layers, making it practical for real-world deployment.
Extensive experiments on real-world mobility datasets demonstrate significant improvements over existing baselines, particularly in challenging real-world settings characterized by spatial noise and dense POI clustering.

\end{abstract}

\keywords{point-of-interest attribution, human mobility modeling, transformers}

\maketitle 

\section{Introduction}
The widespread adoption of mobile devices and location-aware services has dramatically increased the availability of location-based data, 
and a growing body of research leverages mobility data for various downstream geo-spatial tasks, such as next-location prediction \cite{yao2017serm, petzold2006comparison, hsu2024trajgpt, xue2021mobtcast} and detecting anomalies in movement patterns \cite{zhang2024transferable}. A critical first step in these modeling tasks is to contextualize raw GPS trajectories by associating them with semantically-informative user-visited places, or \textit{points of interest (POIs)} -- a process known as \textit{POI attribution} \cite{zheng2011computing,lian2011learning}. 

More specifically, POI attribution refers to the task of associating a user's stay at raw GPS coordinates to a specific location, such as a store, restaurant, or public park. This mapping adds semantic context to otherwise opaque spatial data, enabling more interpretable and actionable insights into human behavior. For example, identifying that a user visited a coffee shop in the afternoon is significantly more informative than simply noting their presence at coordinates $<\text{latitude},\ \text{longitude}>$ at time $t$.

Moreover, effective POI attribution does not merely annotate location data--it enhances the \textit{precision} of spatial understanding. In other words, it improves our ability to interpret and use spatial data accurately \cite{siampoupoly2vec}. Distinguishing which specific establishment a person visited in a multi-store complex (such as a strip mall, for example) yields far more granular and useful behavioral insights than simply knowing they were near the area. This level of precision is vital across various applications, including location-based marketing \cite{cliquet2020location,bauer2016location}, urban planning \cite{natapov2024urban, siampoupoly2vec}, and public health interventions such as pandemic hotspot detection \cite{li2021disparate}. Conversely, misattributing GPS data to the wrong POIs can risk contaminating downstream models to learn misleading or entirely spurious patterns, undermining their reliability and interpretability.

Even though accurate attribution is essential for downstream utility, POI attribution remains underexplored in the literature.
This limited attention is likely due to the inherent challenges of the task. GPS signals are frequently noisy, and in densely populated areas where many POIs are located in close proximity, pinpointing the true destination is non-trivial. For instance, there are 753 retail businesses per square mile in downtown Los Angeles \cite{Shore_2022}, creating significant spatial ambiguity.
Ambiguity becomes unavoidable when the distance between adjacent POIs falls within the 5-15 meter error margin typical of the GPS capabilities of modern smartphones (Table \ref{tab:location_methods}). Alternative positioning methods, such as Wi-Fi or cellular network triangulation, offer broader coverage but at the cost of even lower precision. 

\begin{figure*}[htbp]
  \centering
  \begin{subfigure}[b]{0.48\textwidth}
    \includegraphics[height=5cm]{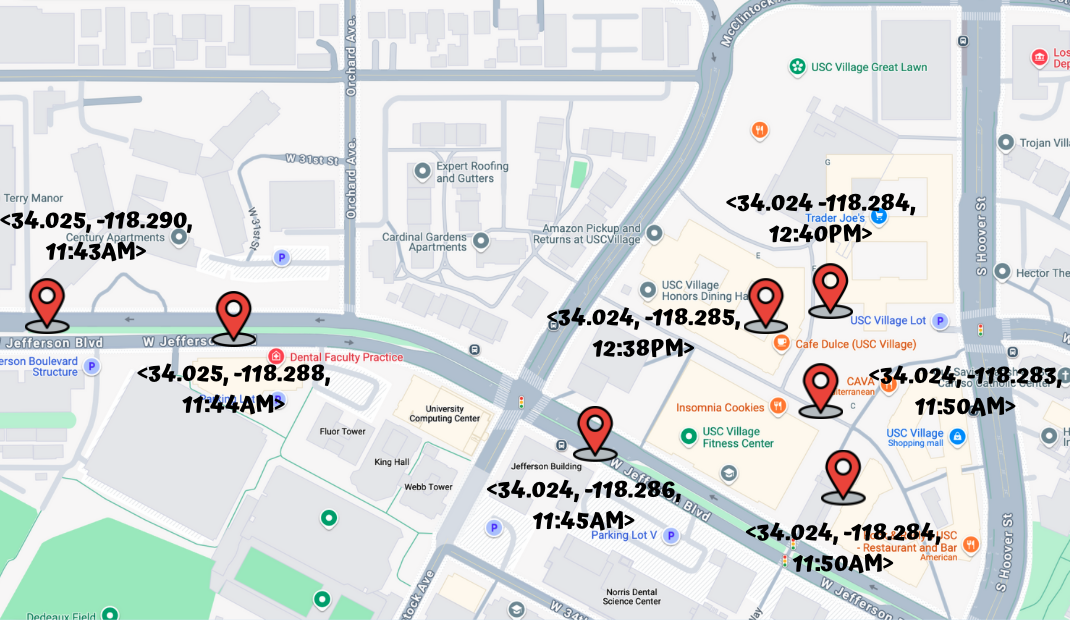}
    \caption{An example of a GPS trace recorded from a mobile device in the format ⟨latitude, longitude, time⟩.}
    \label{fig:1}
  \end{subfigure}
  \hfill
  \begin{subfigure}[b]{0.24\textwidth}
    \includegraphics[height=5cm]{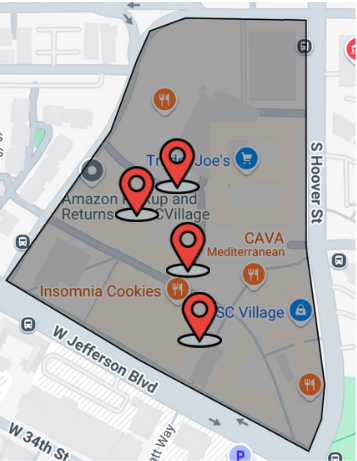}
    \caption{The geographic area identified as a stay is enclosed in black.}
    \label{fig:2}
  \end{subfigure}
  \hfill
  \begin{subfigure}[b]{0.25\textwidth}
    \includegraphics[height=5cm]{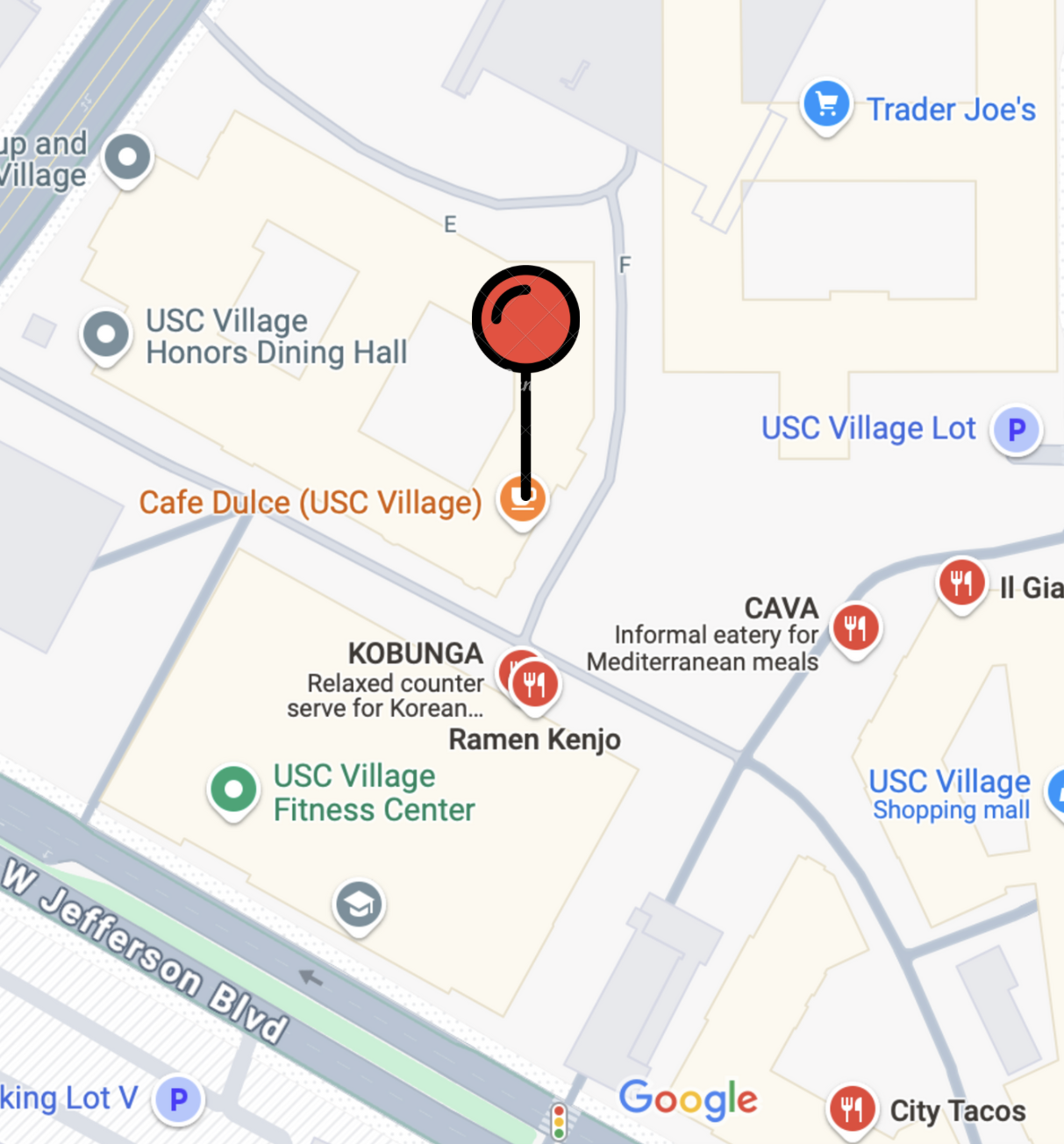}
    \caption{The POI the user visited in the USC Village: Cafe Dulce.}
    \label{fig:3}
  \end{subfigure}
  \caption{Illustration of key concepts in mobility analysis: Figure \ref{fig:1} showcases a GPS trace recorded from an individual's mobile device over time. Figure \ref{fig:2} shows the identification of a stay point based on significant time spent in the area by the individual. Finally, Figure \ref{fig:3} shows the attribution of the stay to a specific POI (Cafe Dulce) within the stay area.}
  \label{fig:three_figs}
\end{figure*}

Yet, despite this complexity, attribution is often reduced to a simple heuristic: assign each stay to the nearest POI \cite{psyllidis2022points}. While this approach is easy to implement, it fails to account for key real-world complexities, including GPS noise and spatial densities that can exceed precision bounds, or to leverage informative contextual cues such as time of day or visit duration. 
For example, consider a stay detected in USC Village, a university campus neighborhood in Los Angeles County. Even if GPS points cluster near Cafe Dulce (Figure \ref{fig:three_figs}), nearby venues--such as a dining hall or clothing store--fall well within the typical error margin of A-GPS or Wi-Fi or cellular positioning (Table \ref{tab:location_methods}). Such scenarios are common in dense urban areas, where POIs are tightly packed together. In these cases, proximity alone is insufficient for reliable attribution. Real-world settings such as this underscore the need for more sophisticated POI attribution models that account for spatial uncertainty and incorporate environmental context and historical user behavior to \textit{predict} the POI visited.

Recent POI attribution methods have moved beyond simple spatial proximity, incorporating signals such as dwell time (i.e., the duration of the stay), user history, or POI semantic information--structured information describing the nature and function of a place (e.g., identifying a location as a cafe, a clinic, or a supermarket, along with attributes like opening hours, typical visitation patterns, or business category). 
For example, Nishida et al. \cite{nishida2014probabilistic} propose a hierarchical Bayesian model that leverages semantic information from POI categories--useful for making inferences into what types of POI a user is more likely to visit--along with individual-level behavioral signals. 
Suzuki et al. \cite{suzuki2019personalized} develop a privacy-preserving approach that attributes visits based solely on a single individual's mobility history, without aggregating across users. 
Finally, the current state-of-the-art, developed by the data company SafeGraph \cite{safegraph_2025}, enriches attribution with detailed spatial features like building footprint polygons and spatial hierarchy metadata that capture spatial relationships among locations. For instance, the hierarchical metadata can indicate that a store is located within a mall or that a food outlet is situated in a food court, providing valuable contextual information that supports more accurate POI attribution.
Collectively, these methods demonstrate the growing recognition that POI attribution benefits from incorporating POI semantics alongside behavioral and contextual signals.
However, despite this progress, most of these approaches face limitations including reliance on unrealistic assumptions, computationally intensive frameworks that limit scalability, and missed opportunities to leverage additional contextual signals. 
For example, Nishida et al. \cite{nishida2014probabilistic} assume that each POI belongs to a single category, which overlooks the multi-functional nature of many real-world locations, such as transit hubs that also serve as commercial centers. Moreover, while their model incorporates a POI's visitation pattern as part of its semantic information, it does not explicitly leverage population-level mobility behaviors. As demonstrated in Section \ref{sec:experiments}, these limitations collectively limit its accuracy. While the approach by Suzuki et al. \cite{suzuki2019personalized} meets requirements of some privacy-preserving frameworks, their reliance on integer linear programming introduces prohibitive computational costs, limiting scalability and utility in non-restrictive settings. While the SafeGraph approach incorporates more detailed spatial contextual information, it omits information from behavioral signals such as individual visit history, population-level patterns, and personal preferences that can be leveraged for more accurate attribution. Furthermore, the comprehensive spatial data layers required by the SafeGraph approach are often unavailable or incomplete in many POI datasets and regions of the world, limiting its generalizability.
While these methods demonstrate the growing recognition of the value of incorporating novel features such as POI semantics, none effectively leverage population-level behavioral mobility patterns that can enhance attribution accuracy in real-world environments.

To address these limitations, we propose \textsf{POIFormer}, a novel Transformer-based framework for POI attribution that jointly models a diverse set of signals, including spatial proximity, temporal features of the visit (arrival/departure and dwell time), POI semantics, user-specific mobility patterns, and population-level historical trends. A key innovation of \textsf{POIFormer} is its explicit incorporation of two dimensions of behavioral context: one capturing individual preferences, and another capturing crowd-level visit patterns. 
Individual preferences are modeled using a transformer that considers both past and future visits, with the location of the current (target) visit masked. This context enables the transformer to evaluate which nearby POI candidate is most likely given a user's past and future visits, based on the time of day and duration of the stay of the target visit. 
Crowd-level historical visit patterns are modeled using the temporal popularity distributions of POIs, estimated via Kernel Density Estimation (KDE).
These KDE models capture the joint distribution of location and time (e.g., hour of day) for visits within each POI category. This enables \textsf{POIFormer} to probabilistically downweight unlikely POIs--for example, reducing the likelihood of assigning a late-night visit to a coffee shop if historical data shows it is rarely  visited at that hour. 
These KDEs are pre-computed per category 
facilitating efficient, scalable inference without sacrificing accuracy since they retain the full joint distribution of location and time while avoiding the need for computation at time of inference. Finally, \textsf{POIFormer} combines individual and crowd-level scores into a unified likelihood measure, selecting the most probable POI (or set of POI) among nearby candidates.
Furthermore, unlike prior approaches \cite{nishida2014probabilistic,safegraph_2025}, \textsf{POIFormer} makes no restrictive assumptions about POI categories, and does not rely on detailed spatial data layers about POIs, thereby enhancing its applicability across diverse geographic and data-constrained contexts. Extensive experimental evaluation on publicly available datasets, one simulated and one derived from real-world mobility traces, demonstrate that \textsf{POIFormer} consistently outperforms existing baselines including the current state-of-the-art technique proposed by SafeGraph \cite{safegraph_2025} by a substantial margin, particularly in top-3 and top-5 accuracy.

The remainder of this paper is organized as follows. Section \ref{sec:related_work} reviews relevant prior work on POI attribution. Section \ref{sec:poiformer} introduces our proposed method, \textsf{POIFormer}, detailing its architecture and components. Experimental results and evaluation are presented in Section \ref{sec:experiments}. Finally, Section \ref{sec:conclusion} concludes the paper and outlines directions for future research.

\rowcolors{2}{gray!10}{white}
\begin{table*}[ht]
\centering
\caption{Comparison of Location Positioning Methods}
\renewcommand{\arraystretch}{1.4}
\begin{tabular}{@{}>{\centering\arraybackslash}p{2.7cm} p{4.5cm} p{2cm} p{3cm} p{4.2cm}@{}}
\toprule
\textbf{Method} & \textbf{How It Works} & \textbf{Avg. Accuracy Indoors / Urban} & \textbf{Avg. Accuracy Outdoors / Rural} & \textbf{Coverage and Best Use Case} \\
\midrule

\textbf{Assisted-GPS (A-GPS)} \newline (commonly used in smartphones) & 
Triangulates signals from multiple satellites via radio waves by measuring signal travel time. Uses cellular assistance to speed up location fixes. &
$\sim$20 m \cite{mok2012initial} &
Modern devices: 7–13 m \cite{GPS_Accuracy,merry2019smartphone} \newline Older devices (such as iPhone 3G 2008): 20–30 m \cite{zandbergen2009accuracy} &
Global. Best for outdoor use in open areas; signal degradation likely in urban canyons or indoors due to obstruction or multipath reflections. \\

\textbf{Handheld GPS Devices} \newline (e.g., Garmin GPSMAP 60Cx) & 
Uses satellite triangulation with advanced antennas and typically supports dual-frequency GPS for improved precision.&
3–5 m \cite{GPS_Accuracy,zandbergen2009accuracy} &
1.6 m (horizontal), 3.1 m (vertical) \cite{zandbergen2009accuracy,GPS_Accuracy} &
High accuracy. Global coverage with good performance in most terrains. \\

\textbf{Wi-Fi Positioning} & 
Estimates location by matching nearby Wi-Fi signals to a known database of access points. &
10–20 m \cite{mok2007location,wallbaum2007priori,swangmuang2008effective,cheong2009gps,zandbergen2009accuracy} &
15–40 m in suburban or semi-urban environments \cite{mok2007location,zandbergen2009accuracy} &
Best in urban or dense environments; limited performance in rural or uncalibrated areas. \\

\textbf{Bluetooth} & 
Estimates location by measuring direction and strength of received signal  & 2-5m
 & up to 700m for direction estimation w/ an antenna array \cite{sollie2022outdoor} &
Susceptible to interference and multipath effects, making it less reliable in complex indoor environments. \\

\textbf{Cellular Network Positioning} & 
Determines location using timing and signal strength from nearby cell towers. &
$\sim$245 m (dense urban) \cite{zandbergen2009accuracy,mohr2008study} &
$\sim$626 m (rural) \cite{zandbergen2009accuracy,mohr2008study} &
Wide coverage. Low precision; best as a fallback method. \\

\textbf{Hybrid (GPS, Wi-Fi, Cellular)} & 
Combines GPS, Wi-Fi, and cellular data to enhance positioning robustness and adaptability. & $\sim$ 4.62m in ideal conditions (i.e., sufficient Wi-Fi and cell tower signals) \cite{fernandez2020hybrid}
 & Accuracy degrades with weaker Wi-Fi and cellular network signals, reverting to 7–13 m \cite{GPS_Accuracy,merry2019smartphone} when relying solely on GPS
 &
Most resilient across environments; balances accuracy and coverage. \\

\bottomrule
\end{tabular}
\label{tab:location_methods}
\end{table*}

\section{Related Work}\label{sec:related_work}
We review prior work related to POI attribution, beginning with stay point detection, which often serves as a first step in the POI-attribution pipeline. Depending on the nature of the data collection process, it may be necessary to first extract stay points from raw GPS trajectories (Figures \ref{fig:1} and \ref{fig:2}) before a POI can be attributed to them (Figure \ref{fig:3}). Thus, we first cover relevant literature on stay point detection, followed by existing work on POI attribution. Lastly, we review work on inferring check-ins from social media data, which is conceptually related to POI attribution but operates without access to trajectory data.

\textbf{Stay Point Detection.}
A fundamental challenge in mobility data analysis is the identification of stay points within raw GPS trajectories. In addition to typical issues such as noisy and imprecise GPS data, there is inherent ambiguity in defining what constitutes a ``stay''--specifically, how long a user must remain stationary for a stop to qualify as a stay point, and how to distinguish intentional stops (for example, dining at a restaurant) from incidental ones (for example, being stuck in traffic). 
At a high level, stay point detection methods can be categorized into clustering-based, grid-based, and stochastic approaches. 

Early work in this area primarily focused on clustering-based methods, including k-means partition-based clustering \cite{zhou2007discovering, cao2010mining, ashbrook2003using} and hierarchical agglomerative clustering \cite{hariharan2004project, krumm2013placer}. These techniques group raw GPS coordinates into clusters representing places and may consider only spatial distance aspects \cite{ester1996density}, or incorporate both spatial and temporal dimensions \cite{zhou2007discovering,palma2008clustering}.
Grid-based or cell-based approaches have also been proposed \cite{agamennoni2009mining, laasonen2004adaptive, grochenig2016cookie}, which divide geographic space into discrete units for analysis. Several stochastic approaches have been proposed as well \cite{scellato2011nextplace, kim2006extracting, venek2016evaluating}, often inferring frequently-visited places as stay points from location data using probabilistic approaches such as Gaussian mixture models \cite{perez2016full,zhang2007adaptive} or Bayesian frameworks \cite{nurmi2008identifying}. 

\textbf{Point of Interest Attribution.} 
A straightforward method frequently employed for POI attribution is the closest centroid approach \cite{cuttone2014inferring}. A stay point is assigned as a visit to the POI whose centroid is closest, provided the distance between the stay point and POI centroid is below a pre-defined threshold. This method performs reasonably well in environments with large, standalone venues such as Walmart or Home Depot, where spatial ambiguity is minimal. However, its effectiveness diminishes significantly in dense urban settings where POIs are closely clustered, or where POIs vary or demonstrate irregularities in size and shape \cite{safegraph_2025}. In such contexts, relying solely on proximity to the centroid often leads to incorrect attributions. These limitations have motivated the development of POI attribution methods that consider additional spatial, temporal, and semantic signals using more sophisticated probabilistic approaches.

One such approach is proposed by Nishida et al. \cite{nishida2014probabilistic}, who introduce a hierarchical Bayesian model for POI attribution. After extracting stay-points from a user's raw GPS trajectory, the model incorporates several factors to infer the most likely POI, including the dwell time, POI category information, and individual user preferences. However, a key limitation of their method is the assumption that each POI is associated with only a single category. This simplification fails to reflect the complexity of real-world locations, where POIs often span multiple categories. For example, a hotel with an in-house restaurant and spa may fall under \textit{lodging}, \textit{restaurant}, and \textit{health and wellness}, while a theatre that also offers dance classes might be categorized as both \textit{entertainment} and \textit{education}. 
The inability to model such multi-category POIs limits the model's applicability in complex urban environments. \textsf{POIFormer} overcomes this limitation by allowing POIs to be represented by multiple categories, enabling a more nuanced and representative attribution in diverse, multi-functional urban settings.

Suzuki et al. \cite{suzuki2019personalized} proposed a personalized POI attribution approach under strict privacy constraints, where user visit and location data remains local to their device. As a result, POI assignment for a given user relies solely on their individual historical data without making use of aggregated information from other users' trajectories and behaviors. 
Their method follows a two-step process. First, stay points are extracted using a variant of conventional methods. Second, POI attribution is formulated as an integer linear programming problem. This optimization considers spatial proximity of the stay point to POIs, dwell time at that stay point, and user-specific behavior patterns such as the user's past visits to various POIs. The formulation of the problem as an integer linear programming, however, leads to extremely high computational costs which limits the practical efficacy of this method. \textsf{POIFormer} maintains computational efficiency by avoiding complexity of optimization formulation with a streamlined Transformer-based architecture that avoids complex optimization, and efficiently incorporates population-level behavioral patterns through pre-computed group-level KDEs that are readily computable from standard mobility data.

The methodology proposed by SafeGraph \cite{safegraph_2025} represents the current state of the art in POI attribution and employs a comprehensive, multi-step pipeline. Beginning with raw GPS trajectories, the approach first filters out implausible GPS signals and clusters the remaining pings to detect stay points. This is accomplished using a two-pass strategy: the first pass clusters pings within large POIs, and the second pass applies a modified DBSCAN algorithm to the remaining pings to form clusters representing potential visits. These clusters are then geospatially joined with an extensive, proprietary database of building footprint polygons that precisely delineate the physical boundaries of individual POIs. To account for any GPS inaccuracies, a buffer is added around each cluster to ensure clusters are accurately matched to the correct POI. To further refine the attribution, their approach incorporates spatial hierarchy metadata, which outlines parent-child relationships between establishments. This is particularly useful in complex structures like malls, where a store (child) is located within a larger building (parent). Next, it incorporates POI semantic context such as type of business as well as temporal information such as time of day when predicting possible POIs. Finally, a scorecard-based ranking mechanism is used to compare pairs of candidate POIs to determine which POI is the most likely for a particular stay. 
While highly effective, SafeGraph's method depends heavily on their proprietary and richly annotated geospatial data layers of POI polygons and spatial hierarchy metadata -- which is not available for many parts of the world.
Moreover, this approach overlooks valuable behavioral signals at both the individual and population level, which can improve accuracy of POI attribution. \textsf{POIFormer} addresses these limiations by explicitly incorporating individual and aggregate behavioral signals, leading to improved POI attribution accuracy, as is demonstrated in Section \ref{sec:experiments}.

\textbf{Determining check-ins from social media data.}
Xi et al. \cite{xi2019modelling} propose a model, Bi-STDDP, that infers missing POI check-ins by integrating bi-directional spatio-temporal dependencies and users' dynamic preferences. In essence, Bi-STDDP utilizes a user's check-ins on social media to predict POIs that may have been visited but not explicitly recorded--i.e., ``missing,'' check-ins. There is no knowledge of the geographic region or location of the missing check-in; instead, it infers the most likely POI based on the other known check-ins and learned user behavior.    
The model effectively incorporates both global spatial and local temporal information to represent the sequential nature of user movements and their evolving behavioral patterns. 
Specifically, Bi-STDDP first extracts the spatial and local temporal information about POIs to glean the complex spatio-temporal dependence relationships. The target temporal pattern is then fed into a multi-layer neural network along with user and POI information to capture the dynamic user preferences. Finally, the dynamic preferences are transformed into the same space as the dependence relationships in the model. 
Meng et al. \cite{meng2017travel} focus on a related task -- determining the purpose of a trip. The authors propose a dynamic Bayesian network model that integrates GPS trajectories, the functionality and POI popularity in trip end areas, which are learned from social media data, to infer users' trip purposes. Trajectory sequences are then classified into one of eight categories: home, education, shopping, eating out, recreational, personal, work, and transportation.

\section{POI Attribution with \textsf{POIFormer}}\label{sec:poiformer}
In this section we describe our methodology for assigning POIs to user stays in mobility trajectories. We cover relevant terminology, formally define the problem, followed by \textsf{POIFormer}'s approach to leverage spatiotemporal context from user trajectories to \textit{predict} the true POI visited during a recorded stay when multiple POI candidates are plausible.

\subsection{Terminology}
We present relevant terminology before formally defining the task of POI attribution. 

\begin{definition}
A \textit{point of interest (POI)} is a specific, named location characterized by associated attributes, such as a category (e.g., `restaurant' or `coffee shop') and precise geographic coordinates. Each POI, $p$, is associated with a set of semantic categories, $C(p)$, which describe its functional characteristics. We use $P$ to denote the set of all the POIs.
\end{definition}

\begin{definition}\label{def:stay}
A \textit{stay} or \textit{stay point} refers to a geographic region where an individual remains for a continuous duration while engaged in a contextually meaningful activity. Formally, a stay is represented as a tuple \textit{($l$, $t^a$, $t^d$, $p$)} where $l$ denotes the geographic location of the stay, $t^a$ denotes the time the individual arrived at that location for this stay, $t^d$ denotes the time the individual departed, and $p$ denotes the POI visited for the stay. For brevity, we denote time tuples as $t = (t^a, t^d)$ henceforth.
\end{definition}

\begin{definition}
A \textit{trajectory} is a chronologically ordered sequence of temporally-disjoint stays recorded for a single individual. It is denoted by $S = \{s_1, s_2, ..., s_n\}$, where each $s_i$ represents a stay as defined in Definition~\ref{def:stay}.
\end{definition}

\paragraph{Problem Definition: POI Attribution}
In a trajectory $S$, stay $s_i = (l_i, t_i, x) \in S$, will need to be attributed to a location $l_i$. 

The task of POI attribution is to identify the most probable POI $p \in P$ that corresponds to $x$, the actual POI destination visited by the individual during stay $s_i$.

\subsection{Methodology}
In this section, we outline \textsf{POIFormer}, a model that can learn to predict the likelihood that a stay occurred at a POI by leveraging user trajectory and crowd-level historical patterns and stay characteristics such as time of day, category of POI, and duration. 
To improve the accuracy of the task of POI attribution, for a visit $s_i$ in an individual's trajectory $S$, we not only leverage the known attributes of the stay, but also the context provided by the individual's trajectory both before and after visit $s_i$ and the overall spatiotemporal priors of the stays at each POI coming from historical mobility analyses.
On a high level, for a given stay point $s_i$ with location $l_i$ and a set of feasible POIs $P' \subseteq P$, a score is computed for each POI $p \in P'$ representing the likelihood that $p$ is the individual's true destination for stay $s_i$ given the context of the trajectory $S$ and the known attributes of the stay $(l_i, t_i)$. The POI with the highest score is considered the attributed POI for that stay.

\subsubsection{Capturing Individual Preferences.}\label{formulation}
For a given stay $s_i$ in a trajectory sequence $S$ and the set of feasible POIs $P' \subseteq P$ near the stay location $l_i$, we aim to learn a model that estimates the probability of each POI being the true destination for that stay, $\Pr(p | s_i, S)$. This captures various signals: the spatial proximity and spatiotemporal characteristics of the stay such as visit time and duration, and the user's visit history. 
For clarity, we denote this going forward as $\Pr(p | t, l, H)$ where $(t, l)$ represents the given stay's spatiotemporal attributes (arrival time, departure time, and location), and $H$ represents the historical and future context surrounding the given stay $s_i$ within the trajectory $S$. Specifically, $H$ includes the sequence of POIs visited before and after $s_i$, as well as their associated semantic categories. This provides the model with insight to behavioral patterns, such as whether the user typically transitions from a gym to a smoothie shop, or from work to a grocery store, which can help disambiguate between candidates in $P'$.

We model the likelihood of a POI $p \in P'$ being the true destination for a given stay $s_i$ as a function of its associated semantic categories.  Each POI $p$ has a set of associated categories,  \( \mathit{C}(p) = \{c_1, c_2, \dots, c_m\} \). With conditional independence assumption among the associated categories, the probability of $p$ being the destination can be approximated as the product of individual probabilities of its constituent categories conditioned on the context of the stay and trajectory sequence. This enables several key advantages: it allows the model to scale efficiently to vast geographic regions with a diverse set of POIs; and it provides robustness in regions where data is sparse, since the estimation of individual category probabilities is more tractable and generalizable than modeling joint category distributions across all POIs. Mathematically, this is represented as:

\begin{equation}\label{eq:one}
    \Pr(p | t, l, H) \approx \prod_{c \in C(p)} \Pr(c | t, l, H)
\end{equation}

\noindent On applying Bayes' rule to $\Pr(c | t, l, H)$, we arrive at: 

\begin{equation}\label{eq:two}
    \Pr(c | t, l, H) = \frac{\Pr(t, l|c, H)\Pr(c|H)}{\Pr(t, l|H)}
\end{equation}
Our partial goal is to obtain the category $c$ that maximizes the term on the left. Since the term in the denominator, $\Pr(t, l|H)$ will be the same for all categories $C(p)$ a POI $p$ belongs to, we omit this term. Furthermore, we make a reasonable conditional independence assumption that $\Pr(t, l|c, H) \approx \Pr(t, l|c)$. Thus, we can approximate Equation \ref{eq:two} as:
\begin{equation}
    \Pr(c | t, l, H) \approx \Pr(t, l|c)\Pr(c|H)
\end{equation}

Therefore, the final probability for a candidate POI $p$ is approximately proportional to the product of the category priors given history, and the spatiotemporal likelihoods given the category, across all categories associated with that POI: 

\begin{equation}
    \Pr(p|t, l, H) \approx \prod_{c \in C(p) } \Pr(t, l | c)\Pr(c | H)
\end{equation}

\textsf{POIFormer} estimates two key probabilistic components: the contextual category prior, $\Pr(c|H)$, and the spatiotemporal category likelihood, $\Pr(t, l|c)$. The term $\Pr(t, l | c)$ captures the distributional characteristics of each category in space and time, effectively encoding semantic patterns associated with POI categories. We train the model using a maximal likelihood approach, specifically minimizing the cross-entropy loss between the predicted scores for the candidate POIs and the ground truth label corresponding to the true destination POI.  

\subsubsection{Capturing Crowd Preferences.}\label{crowd}
To incorporate population-level behavioral patterns $\Pr(t, l | c)$ into POI attribution, we model crowd preferences using category-specific Kernel Density Estimators (KDEs). These KDEs capture the joint spatiotemporal distribution of visits for each POI category, reflecting how visit likelihood varies over both location and time. 

Each KDE is trained offline using a large corpus of historical stay data, where the stay attributes are transformed into projected spatial coordinates, $(x, y)$, and time (for example, hour of day). This approach enables us to learn the fine-grained, category-dependent patterns, such as when and where visits to cafes, train stations, or clinics typically occur, effectively modeling crowd-level mobility behavior.

\subsubsection{Model Architecture.} 
\begin{figure*}
\includegraphics[width=\textwidth]{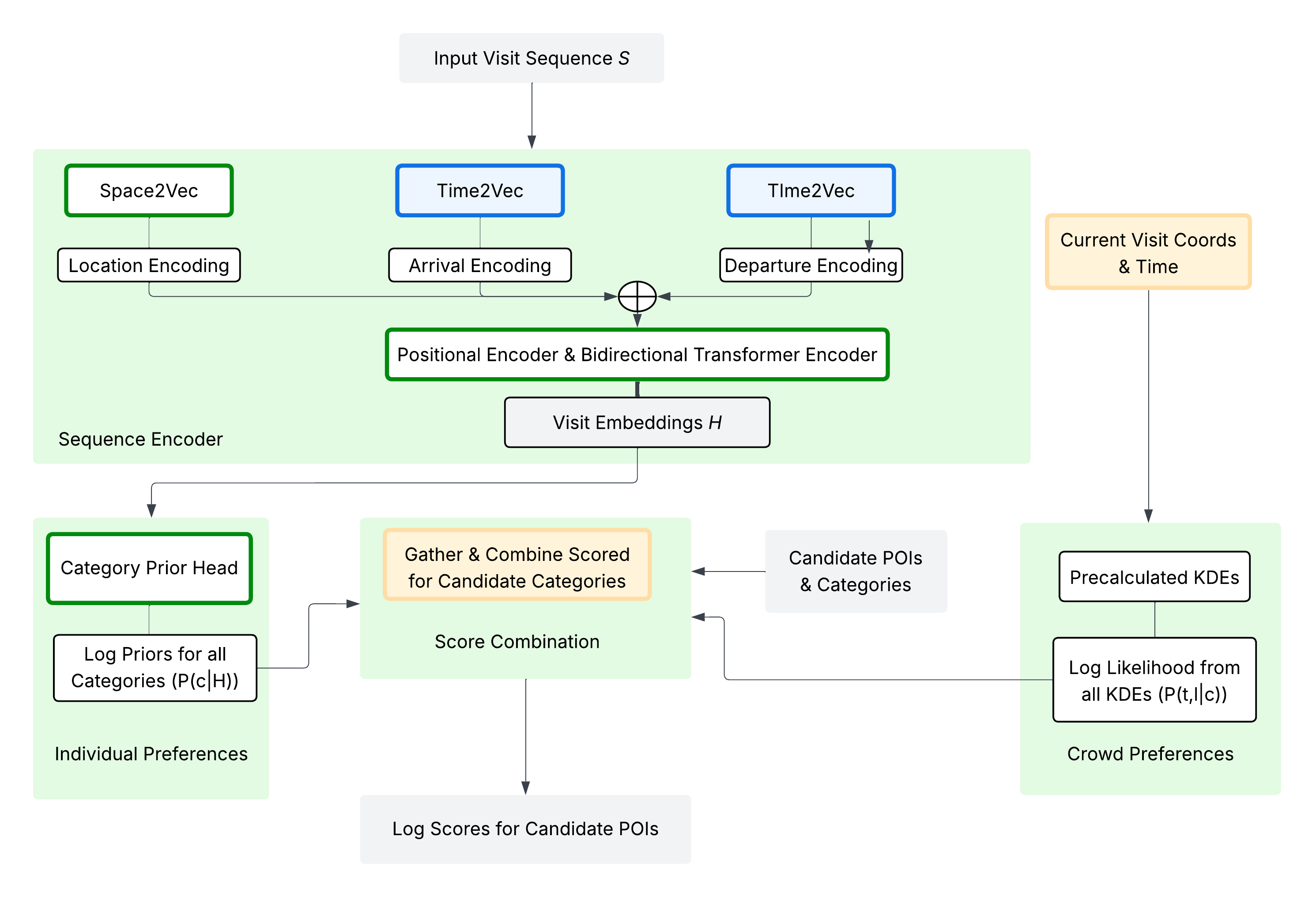}
  \caption{The architecture of \textsf{POIFormer}.}
  \Description{Diagram of the model architecture used in the system.}
  \label{arch}
\end{figure*}

The architecture of \textsf{POIFormer} is illustrated in Figure \ref{arch}. \textsf{POIFormer} consists of several key components. The sequence encoder processes an individual's input trajectory $S$ to capture individual preferences, and a context extractor isolates the relevant contextual embedding corresponding to the stay of interest $s_i$ (Section \ref{sequence}). The context includes both past and future visits, capturing the time and duration of each visit. The location of the missing visit is masked, allowing the model to leverage surrounding visit patterns while inferring the most likely POI at that time.
Following this, the category prior head module predicts the probability distribution over all POI categories based on the extracted context (Section \ref{category}). Meanwhile, the spatiotemporal likelihood module utilizes pre-computed kernel density estimators (KDEs) to evaluate the likelihood of the stay's attributes given a category and the historical visit patterns (Section \ref{spatio}). Finally, the candidate scoring module combines the prior and likelihood information to generate a final score for each candidate POI (Section \ref{scoring}).  

\subsubsection{Sequence Encoding and Context Extraction.}\label{sequence}
To capture the complex spatiotemporal dynamics within a user's trajectory $S$, we first encode each stay $s_i = (l_i, t_i^a, t_i^d, p_i)$ in the sequence. We utilize Space2Vec \cite{mai2020multi} to generate an embedding for the location coordinates $(x_i, y_i)$ derived from location $l_i$, and Time2Vec \cite{kazemi2019time2vec} to generate separate embeddings for the arrival time $t_i^a$ and the departure time $t_i^d$. If $s_i$ is not the stay of interest (i.e., the stay with an unknown POI), we use a learnable embedding matrix to encode $C(p_i)$, the categories of the stay POI. These embeddings for the location, arrival and departure times, and categories if applicable, are subsequently concatenated for each stay to form the input sequence representation. 

This sequence of concatenated embeddings is then processed by a positional encoder to incorporate sequence order information. Finally, a causal transformer encoder, similar to the encoder used by TrajGPT \cite{hsu2024trajgpt}, is used to generate context-aware visit embeddings, $H$. 

Given the full sequence embedding $H$, we extract the specific hidden state $h_i$ corresponding to the stay of interest $s_i$.
This vector $h_i$ represents the learned context for the stay $s_i$ based on the past and future trajectory.

\subsubsection{Category Prior Prediction.}\label{category} 
The extracted context embedding $h_i$ for the stay $s_i$ is used to approximate $\log \Pr(c|H)\ \forall c$, the log prior probability distribution over all possible POI categories. This is achieved using a dedicated linear layer followed by a LogSoftmax activation function:
\begin{equation}
\log \Pr(c|H) := \text{LogSoftmax}(\text{Linear}(h_i)) 
\end{equation}
This output provides a vector where each element represents the log-prior probability of visiting any POI belonging to a specific category, given the trajectory context $H$ ending at $s_i$.

\subsubsection{Spatiotemporal Likelihood Estimation.}\label{spatio}
During model training, the relevant spatiotemporal features of the stay $s_i$ are extracted. To promote robustness, slight gaussian noise is added to coordinates during training. These features are then standardized using a pre-computed scaler that is fitted during KDE training, ensuring consistency across training and inference phases. 

For any given POI category $c$, the 
associated pre-trained KDE is queried to compute the log-likelihood estimate, $\log \Pr(t, l|c)$, of the observed spatiotemporal attributes. Specifically, we evaluate the KDE's log probability density function at the scaled spatiotemporal representation of stay $s_i$, yielding $\log \Pr(t, l|c)$. This likelihood provides a probabilistic measure of how well the stay's spatiotemporal characteristics align with the aggregated behavioral patterns observed for category $c$, allowing the model to downweight unlikely POI candidates accordingly.

\subsubsection{Scoring Candidate POIs .}\label{scoring}
To compute the final score for each candidate POI $p \in P'$, we combine the context-based prior and the spatiotemporal likelihood according to the formulation in Section \ref{formulation}. Since the overall probability is approximately the product $\prod_{c \in C(p)} \Pr(t, l|c) \Pr(c|H)$, we compute the log likelihood by summing the log-probabilities: 

\begin{equation}
   \log \Pr(p|t,l,H) \approx \sum_{c \in C(p)} [\log \Pr(t, l|c) + \log \Pr(c|H)]
\end{equation}

In other words, for each candidate POI $p$, its associated set of categories $C(p)$ is retrieved. For every category $ c \in C(p)$, the log-prior probabilities, $\log \Pr(c|H)$, are obtained from the output of the Category Prior Head (Section \ref{category}). Simultaneously, the Spatiotemporal Likelihood Module (Section \ref{spatio}) is queried to obtain the corresponding log-likelihood estimates $\log \Pr(t, l|c)$. For each category $c \in C(p)$, the log-prior and log-likelihood are summed, $\log \Pr(t, l|c) + \log\Pr(c|H)$. Finally, these combined log-probability terms are aggregated across all categories in $C(p)$ to produce a final score for POI $p$ to arrive at the final score. 

Appropriate masking is applied to handle padding for POIs with fewer than the maximum number of associated categories and to ignore invalid candidates based on the mask. The output is a vector of scores (logits), one for each valid candidate POI in $P'$. 

\subsubsection{Training and Loss.}\label{training}
\textsf{POIFormer} is trained in an end-to-end manner by minimizing the log likelihood loss between the predicted POI distribution and the ground truth label, which specifies the correct POI categories $C(p)$. Formally, the training objective is defined as: 

\begin{align}
     \mathcal{L} &:= -\log \Pr(p|t,l,H) \\
     &\approx -\sum_{c\in C(p)}\left[\log \Pr(t,l|c) + \log \Pr(c|H)\right]
\end{align}

The training process incorporates both positive and negative examples. Optimization is carried out using a stochastic gradient-based optimizer, such as AdamW.

\begin{table*}[ht]
\centering
\caption{Comparison of \textsf{POIFormer} with baseline models for POI Attribution Accuracy on the Breadcrumbs dataset with and without added noise. The best results are highlighted in bold.}
\renewcommand{\arraystretch}{1.4}
 \begin{tabular}{|p{3cm}|>{\centering\arraybackslash}p{1.1cm}|>{\centering\arraybackslash}p{1.1cm}|>{\centering\arraybackslash}p{1.1cm}|
                >{\centering\arraybackslash}p{1.1cm}|>{\centering\arraybackslash}p{1.1cm}|>{\centering\arraybackslash}p{1.1cm}|}
\toprule
\textbf{Method} & \multicolumn{3}{c|}{\textbf{Accuracy w/o Added Noise}} & \multicolumn{3}{c|}{\textbf{Accuracy with Added Noise}} \\
\cline{2-7}
& \textbf{Top-1} & \textbf{Top-3} & \textbf{Top-5} & \textbf{Top-1} & \textbf{Top-3} & \textbf{Top-5} \\
\midrule
Closest centroid \cite{cuttone2014inferring} & 0.8982  & \textbf{1.000} & \textbf{1.000} & 0.6947 & 0.8696 & 0.9032\\
Nishida et al. \cite{nishida2014probabilistic}  &  0.5271 & 0.6749 & 0.6995 & 0.4926 & 0.6305 & 0.6355\\
SafeGraph \cite{safegraph_2025} & \textbf{0.9997} & \textbf{1.000} & \textbf{1.000} &  \textbf{0.9276} & 0.9517 & 0.9655\\
\textbf{\textsf{POIFormer} }& 0.9188 & 0.9924 & \textbf{1.000} & 0.9169 &  \textbf{0.9799} &  \textbf{0.9914} \\
\bottomrule
\end{tabular}
\label{tab:breadcrumbs}
\end{table*}

\begin{table*}[ht]
\centering
\caption{Comparison of \textsf{POIFormer} with baseline models for POI Attribution Accuracy on the NUMOSIM dataset with noise.  \textsf{POIFormer} outperforms all baselines, especially in crowded, densely clustered POI settings—key real-world scenarios that are especially challenging. Best results are highlighted in bold.}
\renewcommand{\arraystretch}{1.4}
\begin{tabular}{|p{4cm}|>{\centering\arraybackslash}p{1.4cm}|>{\centering\arraybackslash}p{1.4cm}|>{\centering\arraybackslash}p{1.4cm}|}
\toprule
\textbf{Method} & \textbf{Top-1 Accuracy} & \textbf{Top-3 Accuracy} & \textbf{Top-5 Accuracy} \\
\midrule
Closest centroid \cite{cuttone2014inferring} & 0.2649 & 0.5911 & 0.7570 \\
Nishida et al. \cite{nishida2014probabilistic} & 0.1002 & 0.1774 & 0.2014 \\
SafeGraph \cite{safegraph_2025} & 0.7023 & 0.7938 & 0.8388 \\
\textbf{\textsf{POIFormer}} & \textbf{0.7223} & \textbf{0.9093} & \textbf{0.9653} \\
\bottomrule
\end{tabular}
\label{tab:numosim}
\end{table*}

\begin{table*}[h]
\centering
\caption{Ablation study of \textsf{POIFormer} components for POI Attribution Accuracy.}
\renewcommand{\arraystretch}{1.4}
\begin{tabular}{|p{5.5cm}|>{\centering\arraybackslash}p{2cm}|>{\centering\arraybackslash}p{2cm}|>{\centering\arraybackslash}p{2cm}|}
\toprule
\textbf{Method} & \textbf{Top-1 Accuracy} & \textbf{Top-3 Accuracy} & \textbf{Top-5 Accuracy} \\
\midrule
\textsf{POIFormer w/o KDE} & 0.6099 &  0.8453 & 0.9183 \\
\textsf{POIFormer w/o learned category prior} & 0.5926  & 0.6894 &  
0.7714\\
\textbf{\textsf{POIFormer} }& \textbf{0.6330} & \textbf{0.8966} & \textbf{0.9514} \\
\bottomrule
\end{tabular}
\label{tab:ablation_table}
\end{table*}

\section{Evaluation}\label{sec:experiments}
We begin by outlining our experimental setup, including the publicly available datasets, evaluation metrics, and baseline methods used for comparison, followed by presentation of experimental results and an ablation study.

\subsection{Experimental Setup}

\subsubsection{Datasets} We utilize two datasets in our experiments: NUMOSIM and Breadcrumbs. NUMOSIM \cite{stanford2024numosim} is a synthetic dataset designed to simulate human mobility patterns in an urban environment, specifically modeled on the city of Los Angeles, California. It is generated using deep generative models trained on real-world travel survey data from the National Household Travel Survey and empirical mobility data \cite{FHWA2022NHTS}. The dataset comprises mobility trajectories for 200,000 agents across two distinct four-week periods (one for training and one for testing), resulting in over 16 million stays. Each stay is labeled with the associated POI and corresponding POI categories. In contrast, Breadcrumbs \cite{moro2019breadcrumbs} is a real-world mobility dataset collected in 2018 in Lausanne, Switzerland, specifically in an area surrounding Lake Geneva--a region with relatively low POI density compared to the highly built-up urban landscape of Los Angeles. It captures mobility data from 81 participants over a three-month period, collected via multiple smartphone sensors, including GPS, WiFI, and Bluetooth. The dataset is validated through detailed participant questionnaires, and stays are also annotated with POIs and POI categories. A comparative summary of the characteristics of both datasets is detailed in Table \ref{tab:dataset_comparison}.

\subsubsection{Processing} Both datasets include pre-identified staypoints. To ensure temporal consistency, we normalize all arrival and departure timestamps by subtracting the earliest arrival time in each dataset from all other timestamps, and then convert the result into seconds. To simulate GPS uncertainty and to enhance robustness and generalizability, we perturb the true geographic coordinates in both datasets by adding zero-mean Gaussian noise. The noise standard deviation ($ \sigma$) is randomly selected from the set \{0.0002, 0.0001, and 0.00005\}. We do not observe significant differences in results across different $\sigma$ values. The NUMOSIM dataset is simulated; therefore, the coordinates of each staypoint exactly match those of its corresponding POI. We add noise to better reflect real-world conditions, where such precision is rare. The Breadcrumbs dataset is based on real-world data collected from smartphones; while some inherent noise is already present (see Table \ref{tab:location_methods} for average error) and the staypoint coordinates do not necessarily coincide with the true annotated POI coordinates, we still add additional noise to increase the challenge of the task, given that the dataset covers a relatively sparse geographic region.

\subsubsection{Metrics} We evaluate POI attribution performance using top-$k$ accuracy. In other words, we measure the proportion of instances where the correct POI appears within a model's top-$k$ ranked predictions, offering a practical view of how well the model prioritizes relevant candidates. We report results for top-1, top-3, and top-5 accuracy to illustrate performance across varying levels of tolerance for prediction rank. Top-1 accuracy denotes the percentage of staypoints in the test set where the model's top predicted POI matches the true POI. Top-3 and top-5 accuracies indicate the percentage of stay points in the test set for which the true POI appears among the model's top 3 or top 5 predicted POI, respectively.

\subsubsection{Baselines} We compare \textsf{POIFormer} with the following: 
\begin{itemize}
    \item The Closest Centroid method \cite{cuttone2014inferring}, a distance-based approach that assigns each staypoint to the nearest POI centroid.
    \item The probabilistic attribution model proposed by Nishida et al. \cite{nishida2014probabilistic}, which models POI assignment as a probabilistic inference problem incorporating spatial and temporal features.
    \item The current state-of-the-art method provided by SafeGraph \cite{safegraph_2025}, which integrates detailed spatial layers such as building polygons and spatial hierarchy metadata with POI features to produce high-accuracy POI attribution. 
\end{itemize}

\subsubsection{Experiments} 
We conduct two sets of experiments. The first set  compares \textsf{POIFormer} with baseline methods on the Breadcrumbs without any added noise. Since the coordinates of each staypoint in the NUMOSIM dataset exactly match those of its corresponding POI, we do not report results for NUMOSIM under noise-free conditions.
The second set of experiments evaluates \textsf{POIFormer} against the same baselines on both Breadcrumbs and NUMOSIM with noise introduced. In each experiment, the goal is to predict (infer) the corresponding POI for each staypoint in the test set. An ``experiment'' in this context refers to evaluating a given method on one dataset, under one noise condition (with or without noise), and for one accuracy metric. In total, we conduct nine experiments: three experiments (Top-1, Top-3, and Top-5 accuracies) on the Breadcrumbs dataset without noise, and three experiments (Top-1, Top-3, and Top-5 accuracies) each for both datasets with added noise.

\subsection{Experimental Results}

\subsubsection{Experiments with added noise.}
As demonstrated in Tables \ref{tab:breadcrumbs} and \ref{tab:numosim}, \textsf{POIFormer} outperforms all baselines in five out of six noisy-data experiments across the Breadcrumbs and NUMOSIM datasets. This highlights \textsf{POIFormer}'s strong robustness to noise, a critical requirement for practical real-world POI attribution systems, given the inherent noise and limits to precision in GPS mobility data.
Notably, \textsf{POIFormer} consistently delivers superior performance in the scenarios that matter most for real-world applications: crowded environments with densely clustered POIs, which are notoriously difficult to handle effectively.

We examine in more detail the single instance where SafeGraph outperforms \textsf{POIFormer} (by 1.15\% in Top-1 accuracy): the Breadcrumbs dataset under noisy conditions. All methods achieve relatively high performance on this dataset, which is expected given its context. The data were collected from 81 users in Lausanne, Switzerland--a small, low-density town of approximately 140,000 residents. Such a setting presents a limited number of POI candidates, reduced spatial ambiguity, and lower mobility diversity, all of which simplify the attribution problem and naturally lead to higher performance. In this environment, the relatively small dataset size and low number of users do not fully invoke \textsf{POIFormer}'s ability to leverage individual preferences and crowd-level behavioral trends--two key strengths of the model. However, despite that, \textsf{POIFormer} nonetheless achieves a highly competitive Top-1 accuracy of 91.69\%, trailing SafeGraph by only 1.15\%. This result further underscores the model's versatility: even in settings where its full modeling capacity is underutilized, \textsf{POIFormer} delivers strong, competitive performance.

Across both datasets, the advantage of \textsf{POIFormer} is even more pronounced for Top-3 and Top-5 accuracy levels.
On NUMOSIM, \textsf{POIFormer} surpasses the second-best method (SafeGraph) by 14.55\% in Top-3 accuracy and by 15.08\% in Top-5 accuracy. On the Breadcrumbs dataset, it outperforms SafeGraph by 2.96\% in Top-3 accuracy and by 2.68\% in Top-5 accuracy, demonstrating its consistent ability to maintain high-quality attribution across diverse and noisy environments.

These results underscore a key strength of \textsf{POIFormer}: it maintains robust and stable performance across varying degrees of data difficulty and noise. While SafeGraph attains near-perfect accuracy under noise-free conditions across both datasets, it exhibits notable degradation under noisy conditions and high-density settings, without recovering performance in Top-3 or Top-5 accuracy as does \textsf{POIFormer}.  
This drop reveals a critical limitation of the state-of-the-art method: while highly effective under clean conditions, it is vulnerable when exposed to the noisy data and dense settings that typify real-world urban environments.

Among all baselines, the Closest Centroid method exhibits the most pronounced performance collapse under noise, underscoring its lack of robustness and its reliance on overly simplistic spatial heuristics. Its relatively strong performance on the Breadcrumbs dataset is unsurprising--Lausanne's sparse and low-density landscape means that spatial proximity alone often suffices for reasonably accurate POI inference. 
However, this heuristic fails to generalize to denser, more complex urban environments like NUMOSIM, where its performance degrades severely, highlighting the need for more sophisticated and noise-resilient approaches such as \textsf{POIFormer}.

The probabilistic model proposed by Nishida et al. \cite{nishida2014probabilistic} performs the worst overall, even trailing the simple Closest Centroid baseline. This is likely due to a key modeling limitation: the assumption that each POI belongs to a single POI category only, forcing an oversimplified representation of POI semantics that can furthermore be inaccurately inferred. Because many POIs are multi-functional, this rigid categorization constrains the model's capacity to capture real-world visitation patterns. Notably, as shown in Table \ref{tab:breadcrumbs} and Table \ref{tab:numosim}, the model performs significantly better on the Breadcrumbs dataset. We hypothesize that this is due in part to the relative simplicity of the Breadcrumbs data, which, in addition to being sparser and lower in density, predominantly contains stays annotated with a single POI category--an artifact that fortuitously aligns with the model's single-category assumption. In contrast, this assumption breaks down in the NUMOSIM dataset, which is substantially denser and where the majority of POIs are associated with three or more categories, leading to markedly poorer performance.

Overall, these findings underscore the importance of modeling spatial and temporal dynamics while capturing both spatial contextual features of POI and behavioral features including observed individual preferences and crowd-level behavioral trends--all critical properties embodied in \textsf{POIFormer}. These additional features likely help to maintain robustness to noise and category ambiguity, evidenced in performance results.

\subsubsection{Experiments without added noise.} In the noise-free setting on the Breadcrumbs dataset, all models achieve relatively high attribution accuracy. The SafeGraph model achieves near-perfect results, with a Top-1 accuracy of 0.9997 and perfect Top-3 and Top-5 accuracies, highlighting its effectiveness when clean location signals are available. The Closest Centroid baseline also performs strongly, achieving 0.8982 Top-1 accuracy and perfect Top-3 and Top-5 accuracy. This is expected, as Lausanne's sparse, low-density landscape allows spatial proximity alone to provide sufficient information for accurate POI inference. In contrast, Nishida et al. performs substantially worse, with a Top-1 accuracy of only 0.5271, highlighting its limitations even under ideal conditions. \textsf{POIFormer} also achieves strong performance in the noise-free setting. Notably, \textsf{POIFormer} exhibits remarkably stable performance across both clean and noisy settings, with only a minimal drop in Top-1 accuracy (from 0.9188 without noise to 0.9169 with noise), while competing methods degrade much more sharply under noise. This indicates that \textsf{POIFormer} learns robust representations that generalize well to realistic conditions where GPS noise is present, ensuring consistent attribution accuracy even when raw location signals are unreliable.

\subsection{Ablation Study}
To evaluate the contributions of the key components of \textsf{POIFormer}, we conducted an ablation study. 
\begin{itemize}
    \item \textsf{POIFormer without KDE}: This variant predicts the POI without leveraging the crowd-level spatiotemporal patterns modeled by the KDE, allowing us to assess the impact of population-level behavioral priors on attribution performance.
    \item \textsf{POIFormer without learned category prior}: This variant predicts the POI without utilizing the learned category prior. Since our full model estimates the likelihood of a POI $p$ being the true destination for a given stay $s_i$ in part as a function of its associated semantic categories, this ablation isolates the contribution of the learned category prior to overall performance.
\end{itemize}

The ablation study is conducted on a subset of the NUMOSIM dataset comprising 100,000 stays. The results, shown in Table \ref{tab:ablation_table}, demonstrate that \textsf{POIFormer} substantially outperforms both ablated variants in POI attribution across all Top-$k$ metrics, underscoring the effectiveness of the proposed modeling approach. 

Removing the KDE component leads to a notable degradation in performance: Top-1 accuracy drops by 2.31\% (from 63.30\% to 60.99\%), Top-3 accuracy decreased by 5.13\%, and Top-5 accuracy falls by 3.31\%. This highlights the important role of KDEs summarizing crowd-level mobility patterns in providing fine-grained spatial and temporal likelihood estimation, which is especially valuable in complex urban settings like NUMOSIM. 

The impact of removing the learned category prior is even more pronounced. Without this component, Top-1 accuracy decreases by 4.04\%, and Top-3 and Top-5 accuracies drop sharply by 20.72\% and 18.00\%, respectively. This result confirms that modeling category-level priors is essential for disambiguating among semantically similar POIs--particularly in high-density environments where multiple POIs may be spatially proximate but serve distinct functions. 

Overall, these findings demonstrate the importance of both components within \textsf{POIFormer}, and demonstrate that jointly modeling spatial, temporal, POI semantics, and behavioral signals is crucial for achieving robust POI attribution performance. The removal of these components results in a substantial decrease in performance, confirming their necessity in capturing the complex spatiotemporal patterns from human mobility data needed for accurate POI attribution in real-world environments.

\section{Conclusion}\label{sec:conclusion}
In this study, we presented \textsf{POIFormer}, a transformer-based model designed to leverage individual preferences and population-level behavioral trends, traditional spatiotemporal signals such as visit and dwell time and POI semantics, to improve POI attribution under noisy and sparse location data generated in dense urban environments. Through comprehensive evaluations on the NUMOSIM and Breadcrumbs datasets, we demonstrated that \textsf{POIFormer} offers substantial improvements in robustness to noise and overall attribution accuracy, particularly in complex urban environments where existing baselines degrade significantly. Moreover, the model demonstrates strong versatility, 
achieving state-of-the-art performance and outperforming nearly all baselines--most notably in Top-3 and Top-5 accuracy. 
In particular, \textsf{POIFormer} consistently outperforms all baselines in the scenarios that matter most for real-world applications—crowded areas with densely clustered POIs—which are notoriously challenging to handle effectively.
This strong performance persists even in settings where its full modeling capacity is underutilized, i.e. low-density urban settings.  
Given its ability to model both individual and population-level behavioral context together with POI semantics, \textsf{POIFormer} can be readily applied to improve POI attribution in a wide range of downstream mobility analytics tasks, including personalized recommendations, urban planning, and public health exposure studies where capturing nuanced spatiotemporal and behavioral patterns is useful for improving predictive accuracy and interpretability.

\bibliographystyle{ACM-Reference-Format}
\bibliography{ref}


\begin{thebibliography}{52}


\ifx \showCODEN    \undefined \def \showCODEN     #1{\unskip}     \fi
\ifx \showISBNx    \undefined \def \showISBNx     #1{\unskip}     \fi
\ifx \showISBNxiii \undefined \def \showISBNxiii  #1{\unskip}     \fi
\ifx \showISSN     \undefined \def \showISSN      #1{\unskip}     \fi
\ifx \showLCCN     \undefined \def \showLCCN      #1{\unskip}     \fi
\ifx \shownote     \undefined \def \shownote      #1{#1}          \fi
\ifx \showarticletitle \undefined \def \showarticletitle #1{#1}   \fi
\ifx \showURL      \undefined \def \showURL       {\relax}        \fi
\providecommand\bibfield[2]{#2}
\providecommand\bibinfo[2]{#2}
\providecommand\natexlab[1]{#1}
\providecommand\showeprint[2][]{arXiv:#2}

\bibitem[Agamennoni et~al\mbox{.}(2009)]%
        {agamennoni2009mining}
\bibfield{author}{\bibinfo{person}{Gabriel Agamennoni}, \bibinfo{person}{Juan Nieto}, {and} \bibinfo{person}{Eduardo Nebot}.} \bibinfo{year}{2009}\natexlab{}.
\newblock \showarticletitle{Mining GPS data for extracting significant places}. In \bibinfo{booktitle}{\emph{2009 IEEE International Conference on Robotics and Automation}}. IEEE, \bibinfo{pages}{855--862}.
\newblock


\bibitem[Ashbrook and Starner(2003)]%
        {ashbrook2003using}
\bibfield{author}{\bibinfo{person}{Daniel Ashbrook} {and} \bibinfo{person}{Thad Starner}.} \bibinfo{year}{2003}\natexlab{}.
\newblock \showarticletitle{Using GPS to learn significant locations and predict movement across multiple users}.
\newblock \bibinfo{journal}{\emph{Personal and Ubiquitous computing}}  \bibinfo{volume}{7} (\bibinfo{year}{2003}), \bibinfo{pages}{275--286}.
\newblock


\bibitem[Bauer and Strauss(2016)]%
        {bauer2016location}
\bibfield{author}{\bibinfo{person}{Christine Bauer} {and} \bibinfo{person}{Christine Strauss}.} \bibinfo{year}{2016}\natexlab{}.
\newblock \showarticletitle{Location-based advertising on mobile devices: A literature review and analysis}.
\newblock \bibinfo{journal}{\emph{Management review quarterly}} \bibinfo{volume}{66}, \bibinfo{number}{3} (\bibinfo{year}{2016}), \bibinfo{pages}{159--194}.
\newblock


\bibitem[Cao et~al\mbox{.}(2010)]%
        {cao2010mining}
\bibfield{author}{\bibinfo{person}{Xin Cao}, \bibinfo{person}{Gao Cong}, {and} \bibinfo{person}{Christian~S Jensen}.} \bibinfo{year}{2010}\natexlab{}.
\newblock \showarticletitle{Mining significant semantic locations from GPS data}.
\newblock \bibinfo{journal}{\emph{Proceedings of the VLDB Endowment}} \bibinfo{volume}{3}, \bibinfo{number}{1-2} (\bibinfo{year}{2010}), \bibinfo{pages}{1009--1020}.
\newblock


\bibitem[Cheong et~al\mbox{.}(2009)]%
        {cheong2009gps}
\bibfield{author}{\bibinfo{person}{Joon~Wayn Cheong}, \bibinfo{person}{Binghao Li}, \bibinfo{person}{Andrew~G Dempster}, {and} \bibinfo{person}{Chris Rizos}.} \bibinfo{year}{2009}\natexlab{}.
\newblock \showarticletitle{GPS/WiFi real-time positioning device: An initial outcome}.
\newblock In \bibinfo{booktitle}{\emph{Location Based Services and TeleCartography II: From sensor fusion to context models}}. \bibinfo{publisher}{Springer}, \bibinfo{pages}{439--456}.
\newblock


\bibitem[Cliquet and Baray(2020)]%
        {cliquet2020location}
\bibfield{author}{\bibinfo{person}{G{\'e}rard Cliquet} {and} \bibinfo{person}{J{\'e}r{\^o}me Baray}.} \bibinfo{year}{2020}\natexlab{}.
\newblock \bibinfo{booktitle}{\emph{Location-based marketing: geomarketing and geolocation}}.
\newblock \bibinfo{publisher}{John Wiley \& Sons}.
\newblock


\bibitem[Cuttone et~al\mbox{.}(2014)]%
        {cuttone2014inferring}
\bibfield{author}{\bibinfo{person}{Andrea Cuttone}, \bibinfo{person}{Sune Lehmann}, {and} \bibinfo{person}{Jakob~Eg Larsen}.} \bibinfo{year}{2014}\natexlab{}.
\newblock \showarticletitle{Inferring human mobility from sparse low accuracy mobile sensing data}. In \bibinfo{booktitle}{\emph{Proceedings of the 2014 ACM International Joint Conference on Pervasive and Ubiquitous Computing: Adjunct Publication}}. \bibinfo{pages}{995--1004}.
\newblock


\bibitem[Ester et~al\mbox{.}(1996)]%
        {ester1996density}
\bibfield{author}{\bibinfo{person}{Martin Ester}, \bibinfo{person}{Hans-Peter Kriegel}, \bibinfo{person}{J{\"o}rg Sander}, \bibinfo{person}{Xiaowei Xu}, {et~al\mbox{.}}} \bibinfo{year}{1996}\natexlab{}.
\newblock \showarticletitle{A density-based algorithm for discovering clusters in large spatial databases with noise}. In \bibinfo{booktitle}{\emph{kdd}}, Vol.~\bibinfo{volume}{96}. \bibinfo{pages}{226--231}.
\newblock


\bibitem[{Federal Highway Administration}(2022)]%
        {FHWA2022NHTS}
\bibfield{author}{\bibinfo{person}{{Federal Highway Administration}}.} \bibinfo{year}{2022}\natexlab{}.
\newblock \bibinfo{title}{2022 NextGen National Household Travel Survey Core Data}.
\newblock \bibinfo{howpublished}{\url{https://nhts.ornl.gov}}.
\newblock
\newblock
\shownote{Available online: \url{https://nhts.ornl.gov}}.


\bibitem[Fern{\'a}ndez et~al\mbox{.}(2020)]%
        {fernandez2020hybrid}
\bibfield{author}{\bibinfo{person}{Pedro~J Fern{\'a}ndez}, \bibinfo{person}{Jos{\'e} Santa}, {and} \bibinfo{person}{Antonio~F Skarmeta}.} \bibinfo{year}{2020}\natexlab{}.
\newblock \showarticletitle{Hybrid positioning for smart spaces: proposal and evaluation}.
\newblock \bibinfo{journal}{\emph{Applied Sciences}} \bibinfo{volume}{10}, \bibinfo{number}{12} (\bibinfo{year}{2020}), \bibinfo{pages}{4083}.
\newblock


\bibitem[Gr{\"o}chenig and Schneider(2016)]%
        {grochenig2016cookie}
\bibfield{author}{\bibinfo{person}{Simon Gr{\"o}chenig} {and} \bibinfo{person}{Cornelia Schneider}.} \bibinfo{year}{2016}\natexlab{}.
\newblock \showarticletitle{A Cookie-Cutter Approach for Determining Places and Stays from Movement Data}.
\newblock \bibinfo{journal}{\emph{GI\_Forum}}  \bibinfo{volume}{1} (\bibinfo{year}{2016}), \bibinfo{pages}{53--64}.
\newblock


\bibitem[Hariharan and Toyama(2004)]%
        {hariharan2004project}
\bibfield{author}{\bibinfo{person}{Ramaswamy Hariharan} {and} \bibinfo{person}{Kentaro Toyama}.} \bibinfo{year}{2004}\natexlab{}.
\newblock \showarticletitle{Project Lachesis: parsing and modeling location histories}. In \bibinfo{booktitle}{\emph{International Conference on Geographic Information Science}}. Springer, \bibinfo{pages}{106--124}.
\newblock


\bibitem[Hsu et~al\mbox{.}(2024)]%
        {hsu2024trajgpt}
\bibfield{author}{\bibinfo{person}{Shang-Ling Hsu}, \bibinfo{person}{Emmanuel Tung}, \bibinfo{person}{John Krumm}, \bibinfo{person}{Cyrus Shahabi}, {and} \bibinfo{person}{Khurram Shafique}.} \bibinfo{year}{2024}\natexlab{}.
\newblock \showarticletitle{Trajgpt: Controlled synthetic trajectory generation using a multitask transformer-based spatiotemporal model}. In \bibinfo{booktitle}{\emph{Proceedings of the 32nd ACM International Conference on Advances in Geographic Information Systems}}. \bibinfo{pages}{362--371}.
\newblock


\bibitem[Kazemi et~al\mbox{.}(2019)]%
        {kazemi2019time2vec}
\bibfield{author}{\bibinfo{person}{Seyed~Mehran Kazemi}, \bibinfo{person}{Rishab Goel}, \bibinfo{person}{Sepehr Eghbali}, \bibinfo{person}{Janahan Ramanan}, \bibinfo{person}{Jaspreet Sahota}, \bibinfo{person}{Sanjay Thakur}, \bibinfo{person}{Stella Wu}, \bibinfo{person}{Cathal Smyth}, \bibinfo{person}{Pascal Poupart}, {and} \bibinfo{person}{Marcus Brubaker}.} \bibinfo{year}{2019}\natexlab{}.
\newblock \showarticletitle{Time2vec: Learning a vector representation of time}.
\newblock \bibinfo{journal}{\emph{arXiv preprint arXiv:1907.05321}} (\bibinfo{year}{2019}).
\newblock


\bibitem[Kim et~al\mbox{.}(2006)]%
        {kim2006extracting}
\bibfield{author}{\bibinfo{person}{Minkyong Kim}, \bibinfo{person}{David Kotz}, {and} \bibinfo{person}{Songkuk Kim}.} \bibinfo{year}{2006}\natexlab{}.
\newblock \showarticletitle{Extracting a mobility model from real user traces}.
\newblock  (\bibinfo{year}{2006}).
\newblock


\bibitem[Krumm and Rouhana(2013)]%
        {krumm2013placer}
\bibfield{author}{\bibinfo{person}{John Krumm} {and} \bibinfo{person}{Dany Rouhana}.} \bibinfo{year}{2013}\natexlab{}.
\newblock \showarticletitle{Placer: semantic place labels from diary data}. In \bibinfo{booktitle}{\emph{Proceedings of the 2013 ACM international joint conference on Pervasive and ubiquitous computing}}. \bibinfo{pages}{163--172}.
\newblock


\bibitem[Laasonen et~al\mbox{.}(2004)]%
        {laasonen2004adaptive}
\bibfield{author}{\bibinfo{person}{Kari Laasonen}, \bibinfo{person}{Mika Raento}, {and} \bibinfo{person}{Hannu Toivonen}.} \bibinfo{year}{2004}\natexlab{}.
\newblock \showarticletitle{Adaptive on-device location recognition}. In \bibinfo{booktitle}{\emph{International Conference on Pervasive Computing}}. Springer, \bibinfo{pages}{287--304}.
\newblock


\bibitem[Li et~al\mbox{.}(2021)]%
        {li2021disparate}
\bibfield{author}{\bibinfo{person}{Qingchun Li}, \bibinfo{person}{Liam Bessell}, \bibinfo{person}{Xin Xiao}, \bibinfo{person}{Chao Fan}, \bibinfo{person}{Xinyu Gao}, {and} \bibinfo{person}{Ali Mostafavi}.} \bibinfo{year}{2021}\natexlab{}.
\newblock \showarticletitle{Disparate patterns of movements and visits to points of interest located in urban hotspots across US metropolitan cities during COVID-19}.
\newblock \bibinfo{journal}{\emph{Royal Society open science}} \bibinfo{volume}{8}, \bibinfo{number}{1} (\bibinfo{year}{2021}), \bibinfo{pages}{201209}.
\newblock


\bibitem[Lian and Xie(2011)]%
        {lian2011learning}
\bibfield{author}{\bibinfo{person}{Defu Lian} {and} \bibinfo{person}{Xing Xie}.} \bibinfo{year}{2011}\natexlab{}.
\newblock \showarticletitle{Learning location naming from user check-in histories}. In \bibinfo{booktitle}{\emph{Proceedings of the 19th ACM SIGSPATIAL International Conference on Advances in Geographic Information Systems}}. \bibinfo{pages}{112--121}.
\newblock


\bibitem[Mai et~al\mbox{.}(2020)]%
        {mai2020multi}
\bibfield{author}{\bibinfo{person}{Gengchen Mai}, \bibinfo{person}{Krzysztof Janowicz}, \bibinfo{person}{Bo Yan}, \bibinfo{person}{Rui Zhu}, \bibinfo{person}{Ling Cai}, {and} \bibinfo{person}{Ni Lao}.} \bibinfo{year}{2020}\natexlab{}.
\newblock \showarticletitle{Multi-scale representation learning for spatial feature distributions using grid cells}.
\newblock \bibinfo{journal}{\emph{arXiv preprint arXiv:2003.00824}} (\bibinfo{year}{2020}).
\newblock


\bibitem[Meng et~al\mbox{.}(2017)]%
        {meng2017travel}
\bibfield{author}{\bibinfo{person}{Chuishi Meng}, \bibinfo{person}{Yu Cui}, \bibinfo{person}{Qing He}, \bibinfo{person}{Lu Su}, {and} \bibinfo{person}{Jing Gao}.} \bibinfo{year}{2017}\natexlab{}.
\newblock \showarticletitle{Travel purpose inference with GPS trajectories, POIs, and geo-tagged social media data}. In \bibinfo{booktitle}{\emph{2017 IEEE International Conference on Big Data (Big Data)}}. IEEE, \bibinfo{pages}{1319--1324}.
\newblock


\bibitem[Merry and Bettinger(2019)]%
        {merry2019smartphone}
\bibfield{author}{\bibinfo{person}{Krista Merry} {and} \bibinfo{person}{Pete Bettinger}.} \bibinfo{year}{2019}\natexlab{}.
\newblock \showarticletitle{Smartphone GPS accuracy study in an urban environment}.
\newblock \bibinfo{journal}{\emph{PloS one}} \bibinfo{volume}{14}, \bibinfo{number}{7} (\bibinfo{year}{2019}), \bibinfo{pages}{e0219890}.
\newblock


\bibitem[Mohr et~al\mbox{.}(2008)]%
        {mohr2008study}
\bibfield{author}{\bibinfo{person}{Marian Mohr}, \bibinfo{person}{Christopher Edwards}, {and} \bibinfo{person}{Ben McCarthy}.} \bibinfo{year}{2008}\natexlab{}.
\newblock \showarticletitle{A study of LBS accuracy in the UK and a novel approach to inferring the positioning technology employed}.
\newblock \bibinfo{journal}{\emph{Computer Communications}} \bibinfo{volume}{31}, \bibinfo{number}{6} (\bibinfo{year}{2008}), \bibinfo{pages}{1148--1159}.
\newblock


\bibitem[Mok and Retscher(2007)]%
        {mok2007location}
\bibfield{author}{\bibinfo{person}{Esmond Mok} {and} \bibinfo{person}{G{\"u}nther Retscher}.} \bibinfo{year}{2007}\natexlab{}.
\newblock \showarticletitle{Location determination using WiFi fingerprinting versus WiFi trilateration}.
\newblock \bibinfo{journal}{\emph{Journal of Location Based Services}} \bibinfo{volume}{1}, \bibinfo{number}{2} (\bibinfo{year}{2007}), \bibinfo{pages}{145--159}.
\newblock


\bibitem[Mok et~al\mbox{.}(2012)]%
        {mok2012initial}
\bibfield{author}{\bibinfo{person}{Esmond Mok}, \bibinfo{person}{Guenther Retscher}, {and} \bibinfo{person}{Chen Wen}.} \bibinfo{year}{2012}\natexlab{}.
\newblock \showarticletitle{Initial test on the use of GPS and sensor data of modern smartphones for vehicle tracking in dense high rise environments}. In \bibinfo{booktitle}{\emph{2012 Ubiquitous Positioning, Indoor Navigation, and Location Based Service (UPINLBS)}}. IEEE, \bibinfo{pages}{1--7}.
\newblock


\bibitem[Moro et~al\mbox{.}(2019)]%
        {moro2019breadcrumbs}
\bibfield{author}{\bibinfo{person}{Arielle Moro}, \bibinfo{person}{Vaibhav Kulkarni}, \bibinfo{person}{Pierre-Adrien Ghiringhelli}, \bibinfo{person}{Bertil Chapuis}, \bibinfo{person}{K{\'e}vin Huguenin}, {and} \bibinfo{person}{Beno{\^\i}t Garbinato}.} \bibinfo{year}{2019}\natexlab{}.
\newblock \showarticletitle{Breadcrumbs: a rich mobility dataset with point-of-interest annotations}. In \bibinfo{booktitle}{\emph{Proceedings of the 27th ACM SIGSPATIAL International Conference on Advances in Geographic Information Systems}}. \bibinfo{pages}{508--511}.
\newblock


\bibitem[Natapov et~al\mbox{.}(2024)]%
        {natapov2024urban}
\bibfield{author}{\bibinfo{person}{Asya Natapov}, \bibinfo{person}{Achituv Cohen}, {and} \bibinfo{person}{Sagi Dalyot}.} \bibinfo{year}{2024}\natexlab{}.
\newblock \showarticletitle{Urban planning and design with points of interest and visual perception}.
\newblock \bibinfo{journal}{\emph{Environment and Planning B: Urban Analytics and City Science}} \bibinfo{volume}{51}, \bibinfo{number}{3} (\bibinfo{year}{2024}), \bibinfo{pages}{641--655}.
\newblock


\bibitem[{National Coordination Office for Space-Based Positioning, Navigation, and Timing}(nd)]%
        {GPS_Accuracy}
\bibfield{author}{\bibinfo{person}{{National Coordination Office for Space-Based Positioning, Navigation, and Timing}}.} \bibinfo{year}{n.d.}\natexlab{}.
\newblock \bibinfo{title}{GPS Accuracy}.
\newblock
\urldef\tempurl%
\url{https://www.gps.gov/systems/gps/performance/accuracy/}
\showURL{%
\tempurl}
\newblock
\shownote{Accessed: 2025-05-27}.


\bibitem[Nishida et~al\mbox{.}(2014)]%
        {nishida2014probabilistic}
\bibfield{author}{\bibinfo{person}{Kyosuke Nishida}, \bibinfo{person}{Hiroyuki Toda}, \bibinfo{person}{Takeshi Kurashima}, {and} \bibinfo{person}{Yoshihiko Suhara}.} \bibinfo{year}{2014}\natexlab{}.
\newblock \showarticletitle{Probabilistic identification of visited point-of-interest for personalized automatic check-in}. In \bibinfo{booktitle}{\emph{Proceedings of the 2014 ACM International Joint Conference on Pervasive and Ubiquitous Computing}}. \bibinfo{pages}{631--642}.
\newblock


\bibitem[Nurmi and Bhattacharya(2008)]%
        {nurmi2008identifying}
\bibfield{author}{\bibinfo{person}{Petteri Nurmi} {and} \bibinfo{person}{Sourav Bhattacharya}.} \bibinfo{year}{2008}\natexlab{}.
\newblock \showarticletitle{Identifying meaningful places: The non-parametric way}. In \bibinfo{booktitle}{\emph{International Conference on Pervasive Computing}}. Springer, \bibinfo{pages}{111--127}.
\newblock


\bibitem[Palma et~al\mbox{.}(2008)]%
        {palma2008clustering}
\bibfield{author}{\bibinfo{person}{Andrey~Tietbohl Palma}, \bibinfo{person}{Vania Bogorny}, \bibinfo{person}{Bart Kuijpers}, {and} \bibinfo{person}{Luis~Otavio Alvares}.} \bibinfo{year}{2008}\natexlab{}.
\newblock \showarticletitle{A clustering-based approach for discovering interesting places in trajectories}. In \bibinfo{booktitle}{\emph{Proceedings of the 2008 ACM symposium on Applied computing}}. \bibinfo{pages}{863--868}.
\newblock


\bibitem[P{\'e}rez-Torres et~al\mbox{.}(2016)]%
        {perez2016full}
\bibfield{author}{\bibinfo{person}{Rafael P{\'e}rez-Torres}, \bibinfo{person}{C{\'e}sar Torres-Huitzil}, {and} \bibinfo{person}{Hiram Galeana-Zapi{\'e}n}.} \bibinfo{year}{2016}\natexlab{}.
\newblock \showarticletitle{Full on-device stay points detection in smartphones for location-based mobile applications}.
\newblock \bibinfo{journal}{\emph{Sensors}} \bibinfo{volume}{16}, \bibinfo{number}{10} (\bibinfo{year}{2016}), \bibinfo{pages}{1693}.
\newblock


\bibitem[Petzold et~al\mbox{.}(2006)]%
        {petzold2006comparison}
\bibfield{author}{\bibinfo{person}{Jan Petzold}, \bibinfo{person}{Faruk Bagci}, \bibinfo{person}{Wolfgang Trumler}, {and} \bibinfo{person}{Theo Ungerer}.} \bibinfo{year}{2006}\natexlab{}.
\newblock \showarticletitle{Comparison of different methods for next location prediction}. In \bibinfo{booktitle}{\emph{Euro-Par 2006 Parallel Processing: 12th International Euro-Par Conference, Dresden, Germany, August 28--September 1, 2006. Proceedings 12}}. Springer, \bibinfo{pages}{909--918}.
\newblock


\bibitem[Psyllidis et~al\mbox{.}(2022)]%
        {psyllidis2022points}
\bibfield{author}{\bibinfo{person}{Achilleas Psyllidis}, \bibinfo{person}{Song Gao}, \bibinfo{person}{Yingjie Hu}, \bibinfo{person}{Eun-Kyeong Kim}, \bibinfo{person}{Grant McKenzie}, \bibinfo{person}{Ross Purves}, \bibinfo{person}{May Yuan}, {and} \bibinfo{person}{Clio Andris}.} \bibinfo{year}{2022}\natexlab{}.
\newblock \showarticletitle{Points of Interest (POI): A commentary on the state of the art, challenges, and prospects for the future}.
\newblock \bibinfo{journal}{\emph{Computational urban science}} \bibinfo{volume}{2}, \bibinfo{number}{1} (\bibinfo{year}{2022}), \bibinfo{pages}{20}.
\newblock


\bibitem[{SafeGraph}(2025)]%
        {safegraph_2025}
\bibfield{author}{\bibinfo{person}{{SafeGraph}}.} \bibinfo{year}{2025}\natexlab{}.
\newblock \bibinfo{title}{Determining Points of Interest Visits From Location Data: A Technical Guide to Visit Attribution}.
\newblock \bibinfo{howpublished}{\url{https://www.safegraph.com/guides/visit-attribution-white-paper}}.
\newblock
\newblock
\shownote{Places Data Curated for Accurate Geospatial Analytics}.


\bibitem[Scellato et~al\mbox{.}(2011)]%
        {scellato2011nextplace}
\bibfield{author}{\bibinfo{person}{Salvatore Scellato}, \bibinfo{person}{Mirco Musolesi}, \bibinfo{person}{Cecilia Mascolo}, \bibinfo{person}{Vito Latora}, {and} \bibinfo{person}{Andrew~T Campbell}.} \bibinfo{year}{2011}\natexlab{}.
\newblock \showarticletitle{Nextplace: a spatio-temporal prediction framework for pervasive systems}. In \bibinfo{booktitle}{\emph{Pervasive Computing: 9th International Conference, Pervasive 2011, San Francisco, USA, June 12-15, 2011. Proceedings 9}}. Springer, \bibinfo{pages}{152--169}.
\newblock


\bibitem[Shore(2022)]%
        {Shore_2022}
\bibfield{author}{\bibinfo{person}{Elan Shore}.} \bibinfo{year}{2022}\natexlab{}.
\newblock \bibinfo{title}{5 Standout Stats About Downtown LA}.
\newblock
\urldef\tempurl%
\url{https://downtownla.com/article/value-of-downtowns\#:\~:text=Downtown’s%20outsized%20economic%20contribution%20is,the%20rest%20of%20the%20city}
\showURL{%
\tempurl}


\bibitem[Siampou et~al\mbox{.}(2025)]%
        {siampoupoly2vec}
\bibfield{author}{\bibinfo{person}{Maria~Despoina Siampou}, \bibinfo{person}{Jialiang Li}, \bibinfo{person}{John Krumm}, \bibinfo{person}{Cyrus Shahabi}, {and} \bibinfo{person}{Hua Lu}.} \bibinfo{year}{2025}\natexlab{}.
\newblock \showarticletitle{Poly2Vec: Polymorphic Fourier-Based Encoding of Geospatial Objects for GeoAI Applications}. In \bibinfo{booktitle}{\emph{Forty-second International Conference on Machine Learning}}.
\newblock


\bibitem[Sollie et~al\mbox{.}(2022)]%
        {sollie2022outdoor}
\bibfield{author}{\bibinfo{person}{Martin~L Sollie}, \bibinfo{person}{Kristoffer Gryte}, \bibinfo{person}{Torleiv~H Bryne}, {and} \bibinfo{person}{Tor~Arne Johansen}.} \bibinfo{year}{2022}\natexlab{}.
\newblock \showarticletitle{Outdoor navigation using bluetooth angle-of-arrival measurements}.
\newblock \bibinfo{journal}{\emph{IEEE Access}}  \bibinfo{volume}{10} (\bibinfo{year}{2022}), \bibinfo{pages}{88012--88033}.
\newblock


\bibitem[Stanford et~al\mbox{.}(2024)]%
        {stanford2024numosim}
\bibfield{author}{\bibinfo{person}{Chris Stanford}, \bibinfo{person}{Suman Adari}, \bibinfo{person}{Xishun Liao}, \bibinfo{person}{Yueshuai He}, \bibinfo{person}{Qinhua Jiang}, \bibinfo{person}{Chenchen Kuai}, \bibinfo{person}{Jiaqi Ma}, \bibinfo{person}{Emmanuel Tung}, \bibinfo{person}{Yinlong Qian}, \bibinfo{person}{Lingyi Zhao}, {et~al\mbox{.}}} \bibinfo{year}{2024}\natexlab{}.
\newblock \showarticletitle{NUMOSIM: A Synthetic Mobility Dataset with Anomaly Detection Benchmarks}. In \bibinfo{booktitle}{\emph{Proceedings of the 1st ACM SIGSPATIAL International Workshop on Geospatial Anomaly Detection}}. \bibinfo{pages}{68--78}.
\newblock


\bibitem[Suzuki et~al\mbox{.}(2019)]%
        {suzuki2019personalized}
\bibfield{author}{\bibinfo{person}{Jun Suzuki}, \bibinfo{person}{Yoshihiko Suhara}, \bibinfo{person}{Hiroyuki Toda}, {and} \bibinfo{person}{Kyosuke Nishida}.} \bibinfo{year}{2019}\natexlab{}.
\newblock \showarticletitle{Personalized visited-poi assignment to individual raw GPS trajectories}.
\newblock \bibinfo{journal}{\emph{ACM Transactions on Spatial Algorithms and Systems (TSAS)}} \bibinfo{volume}{5}, \bibinfo{number}{3} (\bibinfo{year}{2019}), \bibinfo{pages}{1--28}.
\newblock


\bibitem[Swangmuang and Krishnamurthy(2008)]%
        {swangmuang2008effective}
\bibfield{author}{\bibinfo{person}{Nattapong Swangmuang} {and} \bibinfo{person}{Prashant Krishnamurthy}.} \bibinfo{year}{2008}\natexlab{}.
\newblock \showarticletitle{An effective location fingerprint model for wireless indoor localization}.
\newblock \bibinfo{journal}{\emph{Pervasive and Mobile Computing}} \bibinfo{volume}{4}, \bibinfo{number}{6} (\bibinfo{year}{2008}), \bibinfo{pages}{836--850}.
\newblock


\bibitem[Venek et~al\mbox{.}(2016)]%
        {venek2016evaluating}
\bibfield{author}{\bibinfo{person}{Verena Venek}, \bibinfo{person}{Richard Brunauer}, {and} \bibinfo{person}{Cornelia Schneider}.} \bibinfo{year}{2016}\natexlab{}.
\newblock \showarticletitle{Evaluating the Brownian bridge movement model to determine regularities of people’s movements}.
\newblock \bibinfo{journal}{\emph{GI\_Forum}}  \bibinfo{volume}{2} (\bibinfo{year}{2016}), \bibinfo{pages}{20--35}.
\newblock


\bibitem[Wallbaum(2007)]%
        {wallbaum2007priori}
\bibfield{author}{\bibinfo{person}{Michael Wallbaum}.} \bibinfo{year}{2007}\natexlab{}.
\newblock \showarticletitle{A priori error estimates for wireless local area network positioning systems}.
\newblock \bibinfo{journal}{\emph{Pervasive and Mobile Computing}} \bibinfo{volume}{3}, \bibinfo{number}{5} (\bibinfo{year}{2007}), \bibinfo{pages}{560--580}.
\newblock


\bibitem[Xi et~al\mbox{.}(2019)]%
        {xi2019modelling}
\bibfield{author}{\bibinfo{person}{Dongbo Xi}, \bibinfo{person}{Fuzhen Zhuang}, \bibinfo{person}{Yanchi Liu}, \bibinfo{person}{Jingjing Gu}, \bibinfo{person}{Hui Xiong}, {and} \bibinfo{person}{Qing He}.} \bibinfo{year}{2019}\natexlab{}.
\newblock \showarticletitle{Modelling of bi-directional spatio-temporal dependence and users’ dynamic preferences for missing poi check-in identification}. In \bibinfo{booktitle}{\emph{Proceedings of the AAAI conference on artificial intelligence (2019)}}, Vol.~\bibinfo{volume}{33}. \bibinfo{pages}{5458--5465}.
\newblock


\bibitem[Xue et~al\mbox{.}(2021)]%
        {xue2021mobtcast}
\bibfield{author}{\bibinfo{person}{Hao Xue}, \bibinfo{person}{Flora Salim}, \bibinfo{person}{Yongli Ren}, {and} \bibinfo{person}{Nuria Oliver}.} \bibinfo{year}{2021}\natexlab{}.
\newblock \showarticletitle{MobTCast: Leveraging auxiliary trajectory forecasting for human mobility prediction}.
\newblock \bibinfo{journal}{\emph{Advances in Neural Information Processing Systems}}  \bibinfo{volume}{34} (\bibinfo{year}{2021}), \bibinfo{pages}{30380--30391}.
\newblock


\bibitem[Yao et~al\mbox{.}(2017)]%
        {yao2017serm}
\bibfield{author}{\bibinfo{person}{Di Yao}, \bibinfo{person}{Chao Zhang}, \bibinfo{person}{Jianhui Huang}, {and} \bibinfo{person}{Jingping Bi}.} \bibinfo{year}{2017}\natexlab{}.
\newblock \showarticletitle{Serm: A recurrent model for next location prediction in semantic trajectories}. In \bibinfo{booktitle}{\emph{Proceedings of the 2017 ACM on Conference on Information and Knowledge Management}}. \bibinfo{pages}{2411--2414}.
\newblock


\bibitem[Zandbergen(2009)]%
        {zandbergen2009accuracy}
\bibfield{author}{\bibinfo{person}{Paul~A Zandbergen}.} \bibinfo{year}{2009}\natexlab{}.
\newblock \showarticletitle{Accuracy of iPhone locations: A comparison of assisted GPS, WiFi and cellular positioning}.
\newblock \bibinfo{journal}{\emph{Transactions in GIS}}  \bibinfo{volume}{13} (\bibinfo{year}{2009}), \bibinfo{pages}{5--25}.
\newblock


\bibitem[Zhang et~al\mbox{.}(2007)]%
        {zhang2007adaptive}
\bibfield{author}{\bibinfo{person}{Keshu Zhang}, \bibinfo{person}{Haifeng Li}, \bibinfo{person}{Kari Torkkola}, {and} \bibinfo{person}{Mike Gardner}.} \bibinfo{year}{2007}\natexlab{}.
\newblock \showarticletitle{Adaptive learning of semantic locations and routes}. In \bibinfo{booktitle}{\emph{International Symposium on Location-and Context-Awareness}}. Springer, \bibinfo{pages}{193--210}.
\newblock


\bibitem[Zhang et~al\mbox{.}(2024)]%
        {zhang2024transferable}
\bibfield{author}{\bibinfo{person}{Zheng Zhang}, \bibinfo{person}{Hossein Amiri}, \bibinfo{person}{Dazhou Yu}, \bibinfo{person}{Yuntong Hu}, \bibinfo{person}{Liang Zhao}, {and} \bibinfo{person}{Andreas Z{\"u}fle}.} \bibinfo{year}{2024}\natexlab{}.
\newblock \showarticletitle{Transferable Unsupervised Outlier Detection Framework for Human Semantic Trajectories}. In \bibinfo{booktitle}{\emph{Proceedings of the 32nd ACM International Conference on Advances in Geographic Information Systems}}. \bibinfo{pages}{350--360}.
\newblock


\bibitem[Zheng and Zhou(2011)]%
        {zheng2011computing}
\bibfield{author}{\bibinfo{person}{Yu Zheng} {and} \bibinfo{person}{Xiaofang Zhou}.} \bibinfo{year}{2011}\natexlab{}.
\newblock \bibinfo{booktitle}{\emph{Computing with spatial trajectories}}.
\newblock \bibinfo{publisher}{Springer Science \& Business Media}.
\newblock


\bibitem[Zhou et~al\mbox{.}(2007)]%
        {zhou2007discovering}
\bibfield{author}{\bibinfo{person}{Changqing Zhou}, \bibinfo{person}{Dan Frankowski}, \bibinfo{person}{Pamela Ludford}, \bibinfo{person}{Shashi Shekhar}, {and} \bibinfo{person}{Loren Terveen}.} \bibinfo{year}{2007}\natexlab{}.
\newblock \showarticletitle{Discovering personally meaningful places: An interactive clustering approach}.
\newblock \bibinfo{journal}{\emph{ACM Transactions on Information Systems (TOIS)}} \bibinfo{volume}{25}, \bibinfo{number}{3} (\bibinfo{year}{2007}), \bibinfo{pages}{12--es}.
\newblock


\end{thebibliography}

\appendix

\begin{table*}[ht]
\centering
\caption{Comparison of Breadcrumbs and NUMOSIM Datasets}
\renewcommand{\arraystretch}{1.4}
\begin{tabular}{@{}p{3.5cm} p{5.5cm} p{5.5cm}@{}}
\toprule
\textbf{Attribute} & \textbf{Breadcrumbs} & \textbf{NUMOSIM} \\
\midrule
\textbf{Dataset Type} & Real-world mobility dataset & Synthetic mobility dataset with injected anomalies \\
\textbf{Data Collection Method} & Multi-sensor smartphone data (GPS, GSM, Wi-Fi, Bluetooth) from 81 participants & Simulated data using deep learning models trained on real mobility data \\
\textbf{Temporal Coverage} & 12 weeks between March and June 2018 & Simulated over multiple weeks \\
\textbf{Spatial Coverage} & Lausanne, Switzerland & Large-scale simulation of Los Angeles, CA \\
\textbf{Number of Users/Agents} & 81 individuals & 200,000 simulated agents \\
\textbf{Data Modalities} & GPS, GSM, Wi-Fi, Bluetooth, contact records, calendar events, lifestyle information & GPS trajectories, stay points, check-ins, trip durations, trip purposes \\
\textbf{Annotations} & Semantic labels for points of interest (e.g., "home", "work"), demographic attributes, social relationship labels. Stays are annotated with ground truth POIs, POI category confirmed by users. & Annotated anomalies (e.g., centralized manipulation, infectious disease spread), demographic and social attributes. Stays are annotated with POIs and POI categories. \\
\bottomrule
\end{tabular}
\label{tab:dataset_comparison}
\end{table*}

\end{document}